\documentclass[10pt]{article}

\usepackage{amsmath,amssymb}
\usepackage{multirow,array}
\newcolumntype{C}{>{$}c<{$}}
\usepackage{dcolumn}
\usepackage{hyperref} 
\hypersetup{colorlinks=true,
pdfborder={0 0 0},
urlcolor=blue}
\usepackage[inline, shortlabels]{enumitem}
\usepackage{tikz}
\usepackage{float}
\usetikzlibrary{positioning,shapes,arrows,patterns}

\begin{document}
\title{Categorizing Variants of Goodhart's Law}
\author{
  \begin{tabular}[h]{cc}
      David Manheim & Scott Garrabrant \\
      \small{davidmanheim@gmail.com}& \small{scott@intelligence.org}
  \end{tabular} 
}
\date{\today}

\maketitle

There are several distinct failure modes for overoptimization of systems on the basis of metrics. This occurs when a metric which can be used to improve a system is used to such an extent that further optimization is ineffective or harmful, and is sometimes termed Goodhart's Law\footnote{As a historical note, Goodhart’s Law \cite{goodhart} as originally formulated states that ``any observed statistical regularity will tend to collapse once pressure is placed upon it for control purposes.'' This has been interpreted and explained more widely, perhaps to the point where it is ambiguous what the term means. Other closely related formulations, such as Campbell's law (which arguably has scholarly precedence\cite{Rodamar2017}) and the Lucas critique, were also initially specific, and their interpretation has also been expanded greatly. Lastly, the Cobra Effect and perverse incentives are often closely related to these failures, and the different effects interact. Because none of the terms were laid out formally, the categories proposed do not match what was originally discussed. A separate forthcoming paper intends to address the relationship between those formulations and the categories more formally explained here.}. This class of failure is often poorly understood, partly because terminology for discussing them is ambiguous, and partly because discussion using this ambiguous terminology ignores distinctions between different failure modes of this general type.

This paper expands on an earlier discussion by Garrabrant \cite{goodhartcategories}, which notes there are ``(at least) four different mechanisms'' that relate to Goodhart's Law. This paper is intended to explore these mechanisms further, and specify more clearly how they occur. This discussion should be helpful in better understanding these types of failures in economic regulation, in public policy, in machine learning, and in artificial intelligence alignment\cite{superintelligence}. The importance of Goodhart effects depends on the amount of power directed towards optimizing the proxy, and so the increased optimizationpower offered by artificial intelligence makes it especially critical for that field.

\section*{Varieties of Goodhart-like Phenomena}

As used in this paper, a Goodhart effect is when optimization causes a collapse of the statistical relationship between a goal which the optimizer intends and the proxy used for that goal. The four categories of Goodhart effects introduced by Garrabrant are \begin{enumerate*} [1) ]%
\item Regressional, where selection for an imperfect proxy necessarily also selects for noise, \item Extremal, where selection for the metric pushes the state distribution into a region where old relationships no longer hold, \item Causal, where an action on the part of the regulator causes the collapse, and \item Adversarial, where an agent with different goals than the regulator causes the collapse.\end{enumerate*} These varied forms often occur together, but defining them individually is useful. In doing so, this paper introduces and explains several subcategories which differ in important ways.

To formalize the intuitive description, we consider a system $S$ with a set of possible states $s \in S$. For the initial discussion we focus on a single actor, the regulator, who influences the system by selecting a permissable region of the state-space, $A\subseteq S$. For this discussion, we will use $Goal$ to refer to the true goal of the regulator, which is a mapping from states $G(s)\to \mathbb{R}$ for $ s \in S $. Because regulators have incomplete knowledge, they cannot act based on $G(s)$ and instead act only on a proxy $M(s) \to \mathbb{R}$ for $s \in S$\footnote{In general, a mapping from $s\to \mathbb{R}$ is a measaure, and if used for decision-making, it is known as a metric. The current presentation assumes a single-dimensional case. Use of multiple metrics and restrictions follows similar dynamics, but for discussing Goodhart effects the single dimensional case is ideal.}. 
For simplicity's sake, we will consider actions where the regulator chooses some threshold $c$ and the permissible states are defined such that $s \in A$ if $M(s) \geq c$. This creates a selection pressure that allows the first two Goodhart-like effects to occur\footnote{This restriction of the available states is one form of selection pressure. There are other forms of selection pressure which can apply, but these are unnecessary for presenting the basic dynamics. For example, we often find that the states are chosen probabilistically and the distribution can be influenced to prefer certain regions. One important such case is evolutionary selection, where the most likely states are generated based on a set of states selected in a previous generation.}.

\subsection{Regressional Goodhart}

\begin{description}
\item[Regressional Goodhart] - When selecting for a proxy measure, you select not only for the true goal, but also for the difference between the proxy and the goal. This is also known as ``Tails come apart.'' \cite{tails}
\end{description} 
\textbf{Simple Model}:  \begin{equation}M = G + normal(\mu , \sigma ^2) \end{equation} 

\noindent Due to the noise, a point with a large M value will likely have a large G value, but also a large noise value. Thus, when M is large, you can expect G to be predictably smaller than M. Despite the lack of bias, for large values of $c$ the values of G when $M>c$ is expected to be higher than otherwise. While this is the simplest Goodhart effect, it is also the most fundamental: it cannot be avoided. No matter what measure is chosen for optimization, an inexact metric necessarily leads to a divergence between the goal and the metric in the tail. 

\subsection{Extremal Goodhart}

\begin{description}
\item[Extremal Goodhart] - Worlds in which the proxy takes an extreme value may be very different from the ordinary worlds in which the relationship between the proxy and the goal was observed. A form of this occurs occurs in statistics and machine learning as ``out of sample prediction.''
\end{description} 

\noindent Extremal Goodhart can occur in two ways. First, the proxy may have been simplified based on either an insufficient number of observations to uncover the true relationship, or because a simpler measure was preferred. In either case, model insufficiency can lead to extremal Goodhart. Secondly, the measure may be a correct estimator only in certain regions of the space. In regions where the generating process differs, the use of the measure leads to extremal Goodhart.

\begin{description}
\item[Extremal Goodhart - Model Insufficiency] - The metric of interest is based on a learned relationship between the goal and the metric which is approximately accurate in the initial region. Selection pressure moves the metric away from the region in which the relationship is most accurate so that the relationship collapses. 
\end{description} 
\textbf{Simple Model}:  \begin{equation} M = G(s_i)+ G'(s_i)  \end{equation}

In the case of model insufficiency, the error is induced by model simplification and inaccuracy, and not a fundamental issue with the inability to build an sufficiently accurate estimator. In the simplest case, model insufficiency involves incomplete understanding of an empirically learned relationship. Even though we assume the relationship is the same within and without the observed range, it is still often more difficult to detect the exact functional relationship based on a limited range. More commonly, model insufficiency occurs because complex systems are approximated using observations of the non-optimized system space.

An example of this phenomenon in public policy is when a relationship is simplified for use as a measure without recognizing how selection pressure applied to the metric makes the simplification problematic. In machine learning this often happens due to underfitting, such as when a relationship is assumed to be a low degree polynomial because higher order polynomial terms are small in the observed region. Selection on the basis of the approximated metric moves towards regions where the higher-order terms are more important, so that use of the machine learning system creates a Goodhart effect.

\begin{description}
\item[Extremal Goodhart - Change in Regime] - The proxy M may be related to G differently in different regions. Even if the correct relationship is learned for the observed region, in the region where the proxy takes an extreme value the relationship to the goal may be fundamentally different. Selection on the basis of the proxy moves into such a region.

\textbf{Simple Model}:  \begin{equation}
    G =
    \begin{cases}
      M + x, & \text{where}\ M <= a \\
      M + y, & \text{where}\ M > a 
    \end{cases}
  \end{equation}

The case of regime change can be due to measurement error so that errors are systemically incorrect in a given region. Alternatively, it can be due to a generating process that differs in defferent regions. An example of the first case is where wind-speed measurements may be systematically biased downwards when the wind-speed exceeds the design tolerances of the instruments. In this case, we expect that the relationship between measured windspeed and wind damage to systematically differ for measured wind-speeds at and above the maximum tolerance. Any selection pressure for the second case would cause entremal Goodhart to occur when a boundary is reached. For example, the relationship between wind speeds and height undergoes a change above the atmospheric boundary layer. Even when the correct relationship is learned for this lower altitude, the relationship changes above it.
\end{description} 

In the above cases of regressional and extremal Goodhart, there are issues with selection pressure for even simple systems when proxies are used. These two classes require selection pressure, but do not involve an intervention on the part of the regulator. When a regulator intervenes in the causal sense, it changes the state-space and allows an additional type of error that depends on the causal structure of the intervention. 

\subsection{Causal Goodhart}
\begin{description}

\item[Causal Goodhart] - When the causal path between the proxy and the goal is indirect, intervening can change the relationship between the measure and proxy. If a regulator intervenes to maximize a metric, the causal pathway can change such that the proxy no longer tracks the goal. In such cases, extreme interventions can be less effective than more moderate ones, and further selection or intervention can be counterproductive.
\end{description} 

\noindent In the earlier cases, the correlational issues found in Goodhart-like effects exist regardless of causal structure. Here, however, a model that ignores causal structure is problematic when the regulator's actions themselves change the relationships. These take the form of Goodhart-like problems, but can be better discussed in terms of the effect of regulator actions on the causal structure. As a simple example, a regulator attempting to use a windspeed model to build windmills must be careful not to ignore the effect the windmills themselves have on wind-speeds. Building many windmills in a given area can alter the wind-speed in the region enough to invalidate the earlier relationship.

Interestingly, this does not require any uncertainty, nor does it require an incorrect understanding of the relationships, unlike earlier cases. Instead, the effect is induced by the regulator's action. The exact way in which this occurs can vary, as discussed below.

\subsection*{Three Causal Goodhart Effects}

There are a three general cases which lead to causal Goodhart, as well as several others that may appear similar, but are actually regressional or extremal Goodhart effects. The below diagram illustrates three classes of causal Goodhart effects. In each, the regulator attempts to act based on a correlation between the measure and the goal by intervening. The node chosen for intervention is shaded.

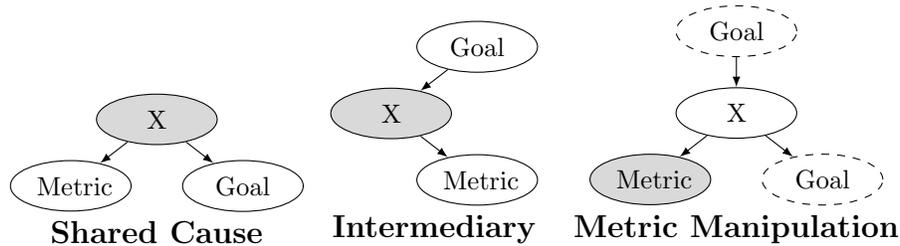
\begin{figure}[htb]
\centering
\begin{tikzpicture}[
  node distance=0.4cm and 0cm,
  mynode/.style={draw,ellipse,text width=0.9cm,align=center}
]
\node[mynode,fill=gray!30] (X) {X};
\node[mynode,below right=of X] (g) {Goal};
\node[mynode,below left=of X] (m) {Metric};
\path (X) edge[-latex] (m);
\path (X) edge[-latex] (g);
\node[below,font=\large\bfseries] at (current bounding box.south) {Shared Cause};
\end{tikzpicture}%
\hspace{0.15cm}
\begin{tikzpicture}[
  node distance=0.4cm and 0cm,
  mynode/.style={draw,ellipse,text width=0.9cm,align=center}
]
\node[mynode] (g) {Goal};
\node[mynode,below left=of g,fill=gray!30] (X) {X};
\node[mynode,below right=of X] (m) {Metric};
\path (g) edge[-latex] (X);
\path (X) edge[-latex] (m);
\node[below,font=\large\bfseries] at (current bounding box.south) {Intermediary};
\end{tikzpicture}%
\hspace{0.15cm}
\begin{tikzpicture}[
  node distance=0.4cm and 0cm,
  mynode/.style={draw,ellipse,text width=0.9cm,align=center}
]
\node[mynode] (X) {X};
\node[mynode,below right=of X,dashed] (g) {Goal};
\node[mynode,above=of X,dashed] (g2) {Goal};
\node[mynode,below left=of X,fill=gray!30] (m) {Metric};
\path (X) edge[-latex] (m);
\path (X) edge[-latex] (g);
\path (g2) edge[-latex] (X);
\node[below,font=\large\bfseries] at (current bounding box.south) {Metric Manipulation};
\end{tikzpicture}%
\end{figure}

\begin{description}
\item[Shared Cause Intervention] - The regulator intervenes on a shared cause of the metric and the goal. 
\end{description}
\textbf{Simple Model}:  The regulator sets the shared cause $X$ to a maximal value. The Metric no longer has a causal relationship to the Goal. The relationship between Metric and Goal now consists entirely of the combination of the error terms between each and $X$.\\

Shared cause intervention may restrict both the Goal and the Metric to a region favored by the regulator, at the cost of changing the metric's relationship to the goal in a way that may be counterproductive for further optimization. Suppose, for example, that there are two tests administered to students which are correlated. If a teacher trains skills related to test taking generally, one of the traits which make the students likely to do well on both tests changes. Because of this, the remaining relationship between scores on the two tests is due to other factors, and the new correlation between scores is likely to be lower than the earlier correlation. This does not imply that the overall outcome is worse than without the optimization, but more extreme interventions can be less effective than less extreme ones. For this reason, further selection or intervention on the basis of the metric can be counterproductive. 

\begin{description}
\item[Intermediary Intervention] - The regulator intervenes on a variable in the causal chain connecting Goal to Metric.
\end{description}
\textbf{Simple Model}:  The regulator sets $X$ to a specific value. The relationship between the Goal and Metric now no longer exists; they are independent variables. This does not necessarily affect the Goal at all, but can serve to increase the value of the metric. 

\begin{description}
\item[Metric Manipulation] - The regulator intervenes to set the Metric, without affecting other nodes. (In this case, it does not matter how the goal was related to X, as indicated in the figure.)
\end{description}
\textbf{Simple Model}:  The regulator sets $M$ to a specific value. The relationship between Metric and Goal now no longer holds.\\

Imagine, to give another example, that a teacher changes test scores or grades\footnote{This is assuming the teacher is the regulator - if the teacher is an agent, this is a different case, explored below.}. This doesn't contribute to the goal of learning, it simply changes $M$ so that it is useless (or, if only some scores are changed, less useful) in measuring $G$. 

The reason for causal Goodhart effects can either be due to mistaken or missing understanding of causal effects, or simple lack of foresight. As with extremal Goodhart effects, mistakes about the causal connection can be due to insufficient data, or a change in regime. Importantly, the same mistakes can also lead to extremal or (worsened) correlational Goodhart effects if there is selection instead of intervention. 

\subsection*{Non-Causal Goodhart Effects in Causal Systems}

In the below three cases, the figures show where the regulator mistakes the form of the causal relationship. The dashed lines are the true causal path, and the dotted lines are the assumed ones. If the regulator attempts to select on the basis of the assumed model, it can lead to worsened regressional Goodhart effects or extremal Goodhart effects, but not causal Goodhart effects. 

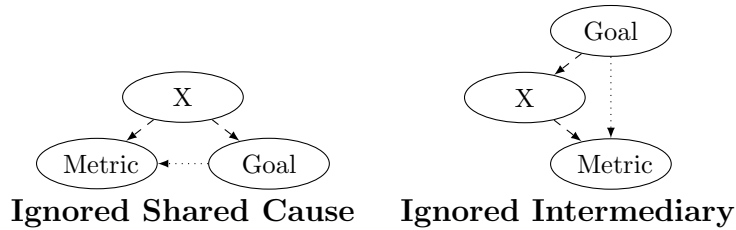
\begin{figure}[htb]
\centering
\begin{tikzpicture}[
  node distance=0.4cm and 0cm,
  mynode/.style={draw,ellipse,text width=0.9cm,align=center}
]
\node[mynode] (X) {X};
\node[mynode,below right=of X] (g) {Goal};
\node[mynode,below left=of X] (m) {Metric};
\path (X) edge[dashed,-latex] (m);
\path (X) edge[dashed,-latex] (g);
\path (g) edge[dotted,-latex] (m);
\node[below,font=\large\bfseries] at (current bounding box.south) {Ignored Shared Cause};
\end{tikzpicture}%
\hspace{0.25cm}
\begin{tikzpicture}[
  node distance=0.4cm and 0cm,
  mynode/.style={draw,ellipse,text width=0.9cm,align=center}
]
\node[mynode] (g) {Goal};
\node[mynode,below left=of g] (X) {X};
\node[mynode,below right=of X] (m) {Metric};
\path (g) edge[dashed,-latex] (X);
\path (X) edge[dashed,-latex] (m);
\path (g) edge[dotted,-latex] (m);
\node[below,font=\large\bfseries] at (current bounding box.south) {Ignored Intermediary};
\end{tikzpicture}%
\end{figure}

\begin{description}
\item[Ignored Shared Cause] - The regulator assumes the relationship between the Goal and Metric is a causal chain, or is direct, but there is instead a shared cause. When ignoring X, the metric is valid as a proxy for the goal. 
\end{description}
\textbf{Simple Model}:  \begin{equation} \begin{split} Metric \thicksim normal(X, \sigma_{m}^2)\\ 
Goal \thicksim normal(X,  \sigma_{g}^2)\end{split}\end{equation} \\

However, because of the conditional independence of Goal and Metric, for any fixed value of $X$ they are uncorrelated.  This means that our metric is less  valid if there is selection. With selection of a fixed X, the case resembles extremal Goodhart, since the regime changed.

\begin{description}
\item[Ignored Intermediary] - The regulator assumes the relationship between the Goal and Metric is direct, but an intermediary exists which creates an additional source of noise.
\end{description}
Without intervention, the causal mistake here adds a term to the error, leading to the earlier cases of regressional and extremal Goodhart effects.

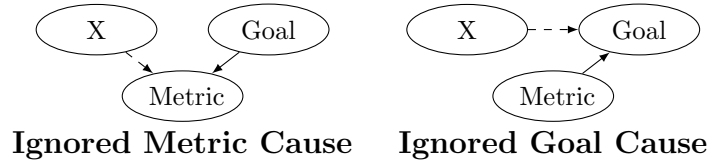
\begin{figure}[htb]
\centering
\begin{tikzpicture}[
  node distance=0.4cm and 0cm,
  mynode/.style={draw,ellipse,text width=0.9cm,align=center}
]
\node[mynode] (g) {Goal};
\node[mynode,below left=of g] (m) {Metric};
\node[mynode,above left=of m] (X) {X};
\path (g) edge[-latex] (m);
\path (X) edge[dashed, -latex] (m);
\node[below,font=\large\bfseries] at (current bounding box.south) {Ignored Metric Cause};
\end{tikzpicture}%
\hspace{0.25cm}
\begin{tikzpicture}[
  node distance=0.4cm and 0cm,
  mynode/.style={draw,ellipse,text width=0.9cm,align=center}
]
\node[mynode] (g) {Goal};
\node[mynode,below left=of g] (m) {Metric};
\node[mynode,above left=of m] (X) {X};
\path (m) edge[-latex] (g);
\path (X) edge[dashed, -latex] (g);
\node[below,font=\large\bfseries] at (current bounding box.south) {Ignored Goal Cause};
\end{tikzpicture}%
\end{figure}

\begin{description}
\item[Ignored Additional Cause] - The metric is caused by multiple factors, of which the goal relates to only some. Alternatively, the goal is caused by multiple factors, of which the metric relates to only some. (No Figure provided.)
\end{description}
\textbf{Simple Model}:  \begin{equation} Metric \thicksim normal(X + Goal, \sigma_{m}^2)\end{equation} \begin{center}or\end{center}  \begin{equation} Goal \thicksim normal(X + Metric, \sigma_{g}^2) \end{equation}
If $X$ has a distribution that matches the distribution of error of the assumed relationship between $X$ and the metric, this leads to worsened regressional Goodhart effects as defined above, because additional noise is due to $X$. In this simple model case the goal is caused by the metric, for example height causing windspeed. More commonly, the metric would be caused by some $Y$, which also causes the Goal. In such a case, the metric might be measured height, where the measurement itself can have error. On the other hand, if $X$ is distributed differently than the assumed distribution of error in the relationship between Goal and Metric, there is a model insufficiency issue. If $X$ contains some nonlinearity, it is a regime change. The earlier examples of extremal Goodhart effects are of these types. (More complex errors can lead to similar mistakes.)

\subsection{Adversarial Goodhart}

The discussion of causal relationships is important for a different class of situations where there are multiple actors. There are numerous such dynamics, but they can be classified as happening in one of two cases. First, the actor may have goals which the regulator is unaware of (or insufficiently wary about) and the agent can act independently of the regulator in a way that adversely affects the regulator's goal. We refer to such cases as adversarial misalignment. Second, the regulator can use incentives to align the agent's goals and use their actions as a way to optimize, while not acting themself. We refer to these as ``Cobra effects.'' In all of these cases, which were called ``adversarial Goodhart'' in the previous work, other agents react in different ways to create Goodhart-like effects following any of the earlier paradigms. We introduce and formulate several specific cases of each.

\begin{description}
\item[Adversarial Misalignment Goodhart] - The agent applies selection pressure knowing the regulator will apply different selection pressure on the basis of the metric\footnote{This is the case most closely related to Campbell's law\cite{campbell}, which was originally stated as  ``The more any quantitative social indicator is used for social decision-making, the more subject it will be to corruption pressures and the more apt it will be to distort and corrupt the social processes it is intended to monitor.'' In this case, the choice of measurement affects the outcome because agents attempt to corrupt the measure.}.\end{description} The adversarial misalignment failure can occur due to the agent creating extremal Goodhart effects, or by exacerbating always-present regressional Goodhart, or due to causal intervention by the agent which changes the effect of the regulator optimization.

\begin{description}
\item[Campbell's Law] - Agents select a metric knowing the choice of regulator metric. Agents can correlate their metric with the regulator's metric, and select on their metric\footnote{They could instead alter causal structure to create the same effect. It seems unclear that this difference is critical to the dynamics considered.}. This further reduces the usefulness of selection using the metric for acheiving the original goal.
\end{description}
\textbf{Simple Model}:  $G_R$, $M_R$ are the regulator's goal and metric. $G_A$, $M_A$ are the agent's goal and metric. (The agent selects $M_A$ after seeing the regulator choice of metric.)
\begin{equation} \begin{split}  M_R = G_R + X \\
 M_A = G_A\cdot X
\end{split}  \end{equation}
Here, the agent selects for values with high $M_A$, and the regulator's later selection then creates a relationship between $X$ and their goal, especially at the extremes. The agent does this by selecting for a metric such that even weak selection on $M_A$ hijacks the regulator's selection on $M_R$ to acheive their goal. The agent choice of metric need not be a useful proxy for their goal absent the regulator's action. In the example given, if $X \thicksim normal(\mu, \sigma^2)$, the correlation between $G_A$ and $M_A$ is zero over the full set of states, but becomes positive on the subspace selected by the regulator.
\\

There are clearly further dynamics worth exploring, but this case serves to introduce the issues involved in adversarial conflict over metrics without incentives. When the regulator attempts to regulate via incentives, a new set of cases can occur, which are generally related to what is know as the ``Cobra Effect.'' \cite{cobra} This is named after a supposed situation in colonial India where British authorities offered a reward for dead cobras. Instead of hunting cobras, however, some people bred and killed their own cobras to kill in order to receive rewards. This not only failed to achieve the goal, but led to more cobras than before the reward was offered. Because of the (supposedly) historical meaning and popular usage, we differentiate between cobra effects that fit this model, which we will call normal cobra effects, and ones where the agent applies pressure in a non-causal fashion to create Goodhart effects, which we call non-causal cobra effects.

\begin{description}

\item[Normal Cobra Effect] - The regulator modifies the agent goal, usually via an incentive, to correlate it with the regulator metric. The agent then acts by changing the observed causal structure due to incompletely aligned goals in a way that creates a Goodhart effect.
\end{description}
\textbf{Simple Model}: The agent finds cause Y (an ``Ignored Additional Cause'') which is unobserved or constant in the structure the regulator initially considered, and the agent changes the value of Y in order to maximize the metric. (This is a form of causal Goodhart using metric manipulation.)

\begin{figure}[htb]
\centering
\begin{tikzpicture}[
  node distance=0.4cm and 0cm,
  mynode/.style={draw,ellipse,text width=0.9cm,align=center}
]
\node[mynode] (g) {Goal};
\node[mynode,below left=of g] (m) {Metric};
\node[mynode,fill=gray!30, above left=of m] (Y) {Y};
\path (g) edge[-latex] (m);
\path (Y) edge[-latex] (m);
\node[below,font=\large\bfseries] at (current bounding box.south) {Cobra Effect};
\end{tikzpicture}
\end{figure}

\begin{equation}  G_{A_R} = G_{A_0} + M_R \end{equation}

The Cobra effect can also occur via an agent action that creates any of the above-mentioned causal Goodhart effects, namely shared cause, intermediary, or metric manipulation.
\begin{description}
\item[Non-Causal Cobra Effect] - The regulator modifies the agent goal to make agent actions aligned with the regulator's metric. Under selection pressure from the agent, extremal Goodhart effects occur or regressional Goodhart effects are worsened.
\end{description}
\textbf{Simple Model}:  $G_R$, $M_R$ are the regulator's goal and metric. $G_{A_0}$ and $G_{A_R}$ are the agent's goal before and after regulator modification.

\begin{equation}  G_{A_R} = G_{A_0} + M_R \end{equation}
When the agent applies selection pressure, it creates a Goodhart effect on the regulator metric.

\subsection*{Conclusion}
This paper represents an attempt to categorize a class of the simple statistical misalignments that occur both in any algorithmic system used for optimization, and in many human systems that rely on metrics for optimization. The dynamics highlighted are hopefully useful to explain many situations of interest in policy design, in machine learning, and in specific questions about AI alignment. In policy, these dynamics are commonly encountered but too-rarely discussed clearly. In machine learning, these errors include extremal Goodhart effects due to using limited data and choosing overly parsimonious models, errors that occur due to myopic consideration of goals, and mistakes that occur when ignoring causality in a system. Finally, in AI alignment, these issues are fundamental to both aligning systems towards a goal, and assuring that the system's metrics do not have perverse effects once the system begins optimizing for them.




\end{document}